# UAS-based Automated Structural Inspection Path Planning via Visual Data Analytics and Optimization

Yuxiang Zhao[1], Benhao Lu[2], Mohamad Alipour[3]

[1]Graduate research assistant, Dept. of Civil and Environmental Engineering, University of Illinois Urbana Champaign. Email: zhao132@illinois.edu
[2]Undergraduate research assistant, Zhejiang University - University of Illinois at Urbana Champaign Institute. Email: benhaol2@illinois.edu
[3]Research Assistant Professor, Dept. of Civil and Environmental Engineering, University of Illinois Urbana Champaign. Email: alipour@illinois.edu

## ABSTRACT

Unmanned Aerial Systems (UAS) have gained significant traction for their application in infrastructure inspections. However, considering the enormous scale and complex nature of infrastructure, automation is essential for improving the efficiency and quality of inspection operations. One of the core problems in this regard is electing an optimal automated flight path that can achieve the mission objectives while minimizing flight time. This paper presents an effective formulation for the path planning problem in the context of structural inspections. Coverage is guaranteed as a constraint to ensure damage detectability and path length is minimized as an objective, thus maximizing efficiency while ensuring inspection quality. A two-stage algorithm is then devised to solve the path planning problem, composed of a genetic algorithm for determining the positions of viewpoints and a greedy algorithm for calculating the poses. A comprehensive sensitivity analysis is conducted to demonstrate the proposed algorithm's effectiveness and range of applicability. Applied examples of the algorithm, including partial space inspection with no-fly zones and focused inspection, are also presented, demonstrating the flexibility of the proposed method to meet real-world structural inspection requirements. In conclusion, the results of this study highlight the feasibility of the proposed approach and establish the groundwork for incorporating automation into UAS-based structural inspection mission planning.

# 1. INTRODUCTION

A significant portion of the essential infrastructure that supports society today was constructed several decades ago, and some have already exceeded their intended lifespan. According to the 2021 America's Infrastructure Report Card [1], among all the 617,000 bridges in the U.S., 42% were constructed more than 50 years ago. Considering the massive infrastructure network and expensive manpower in the U.S., it demands substantial investment in manual maintenance to ensure personal and property safety. Furthermore, given the challenging working conditions in infrastructure inspection, manual labor always carries the risk of accidents, including traffic incidents, falls, and exposure to environmental hazards. Although there are no specific statistics on the fatalities and injuries during inspection tasks, the growing inventory of aging infrastructure increases the risk of accidents during inspections, thus highlighting the importance of ensuring inspection safety. Numerous studies have demonstrated that the implementation of UAS technology contributes to the protection of workers' safety [2-4].

With the development of robotics, Unmanned Aerial Systems (UAS) have emerged as a portable, cost-effective, and efficient solution for inspection [5]. Numerous studies have been conducted to investigate the potential of utilizing UAS for improving the inspection of infrastructure assets, including power lines [6][7], buildings [8-10], railway[11][12], and bridges [13-15]. In [16], a UAS-based end-to-end bridge inspection pipeline has been proposed, which promotes the practical application of autonomous structural inspection with UAS. The application of UAS significantly improves the efficiency and quality of the data collection phase, and computer vision and deep learning techniques have seen a rapid increase in popularity for subsequent automated data processing [17-23].

The studies mentioned above have demonstrated the contribution of UAS to improving inspection techniques. However, further automation in planning and controlling UAS flights is crucial to reduce manual labor and enhance robustness. While deep learning-based data processing methods show promise in improving automated infrastructure inspections, the quality of the resulting inspections is highly sensitive to the selection of UAS viewpoints and poses. Additionally, given the large scale of civil infrastructures and the complexity of surrounding environments, effective path planning is essential for enhancing the practical value of UAS-based inspection. There have been multiple explorations on path planning for infrastructure inspection tasks, and the solutions can be categorized into classical approaches and artificial intelligence (AI) based approaches. Some studies regard the path planning task as a Coverage Path Planning (CPP) problem and solve it with various classical search algorithms such as Dijkstra Algorithm [24], Rapid Random Tree (RRT*) [25], and Probabilistic Roadmap (PRM) [26]. However, given the complex space configurations and strict inspection requirements, the classical search algorithms could be less efficient and may produce only locally optimal solutions [27][28]. Therefore, AI-based algorithms have recently gained rapidly rising tractions due to their advantages, such as a better balance between exploration and exploitation, the ability to avoid locally optimal solutions, and easier parallel programming to improve computational speed. Many studies have attempted to solve path planning problems with various AI-based algorithms, including Ant Colony Optimization [29], Particle Swarm Optimization[30], and Genetic Algorithm [31][32]. However, these works are focused on topics other than structural inspections, including construction progress monitoring, ship inspection, geomagnetic navigation, and general robotic path planning algorithms, leaving a research gap in path planning for structural inspections.

Structural inspection path planning (SIPP) constitutes a specialized form of robotic path planning characterized by its alignment with structural inspection requirements and domain expertise. In contrast with conventional path-planning endeavors, SIPP displays distinctive attributes. Primarily, SIPP is unique due to its intrinsic connection to structural inspection, where neglecting potential damage could lead to member failures and even the collapse of infrastructure, resulting in loss of life and property. Therefore, SIPP has stricter requirements on inspection coverage. Unlike path planning for other tasks, which often involve striking a balance between coverage and efficiency, in SIPP, coverage should be treated as a stringent constraint to ensure the quality and reliability of the inspection results. Conversely, efficiency is regarded as the primary optimization objective. The optimization approach proposed in this study allows us to derive the most optimal path while upholding a predetermined coverage guarantee.

Another distinctive attribute of SIPP is its acknowledgment of varying importance levels within different parts of the infrastructure. Certain elements, like the connections, expansion/construction joints, and beam midspans, can be subject to higher deterioration or load demands and may, therefore, hold greater significance. Consequently, uniform inspection may be inadequate, and SIPP necessitates the allocation of distinct weights to different sections to facilitate damage-guided inspection. These weights can be derived through a variety of methods:

1) Previous Inspection Records: Due to the gradual progression of damage in structures, the damage distribution gleaned from previous inspection records can be utilized to accentuate areas where damage has already manifested and is, therefore, more likely to be present.

2) Structural Analysis: Analytical approaches or numerical simulations such as finite element analysis (FEA) can provide insights into which segments of the structure are critical and exhibit an elevated likelihood of sustaining damage. Leveraging this information, areas of heightened risk can be allocated higher weights during path planning.

3) Inspector Expertise: In instances where conducting FEA might not be cost-effective, inspectors can resort to a user interface to stipulate the importance factor for various elements, drawing upon their domain expertise. By aligning with this importance factor, SIPP can prioritize the predefined crucial elements, directing a greater emphasis on their inspection.

Safety constitutes another pivotal factor that underscores the distinctiveness of SIPP. Infrastructures often feature intricate outlines, thereby presenting many obstacles within the 3D space that must be circumvented during inspection. Furthermore, in scenarios like bridge inspections, the presence of UAS in close vicinity to active traffic poses the risk of distracting drivers, potentially leading to traffic accidents. Thus, incorporating functionality within SIPP that enables the delineation of exclusive zones, such as minimum distance thresholds to infrastructure surfaces or the area directly above a bridge deck, becomes necessary.

In summary, SIPP represents a unique path planning challenge within a complex 3D space, with specialized requirements and constraints, setting it apart from traditional path planning problems. A few studies have initiated research on SIPP. In [33], a rule-based path planning method was proposed based on the recognition of structural members. However, rule-based methods are by their nature applicable and effective for predefined types of infrastructure and are challenging to apply to structures with unique shapes. In [30], the authors used particle swarm optimization to solve SIPP, while the distance from the surface to the drone and the pose of the drone are both predetermined, which vastly reduces the flexibility of path planning and makes the generated paths less optimal. In [34], a genetic algorithm was applied for 5-degree-of-freedom (5DOF) path planning (3D location and camera pose, including pitch and yaw) in building construction monitoring. This study provided an excellent example of task-specific path planning in construction engineering. However, the outline of the inspection object was relatively simple (the exterior surface of a building), and the generated viewpoint space was relatively constrained (less than 1,000 points). As a result, both positions and poses could be discretized into a few levels for simplicity of optimization. However, the 5DOF optimization in a large-scale complex space that requires a fine viewpoint grid can be challenging to converge. In addition, coverage was regarded as an objective instead of a constraint. Thus, there was no guarantee for minimum coverage to ensure the comprehensiveness of the inspection.

Expanding on the literature above, our study focuses on formulating an optimization problem to demonstrate the design requirements and feasibility of SIPP and proposing a comprehensive framework for SIPP that works for different types of infrastructures and tasks. To ensure inspection reliability, we regard coverage as a constraint that must exceed a high threshold. When calculating visibility, we applied constraints on the visible distance and inclination angle so that inspection quality can be ensured. To enable optimization convergence for complex geometries with large viewpoint grids and variable spaces, we propose a two-stage optimization approach consisting of a genetic algorithm for determining viewpoint coordinates and a greedy algorithm for determining poses. This scheme effectively addresses the convergence challenges posed by large variable spaces by breaking the unknowns into two separate and smaller optimization steps. The proposed method allows UAS-based inspection to optimize efficiency while ensuring inspection safety and quality. In short, the primary contributions of this work include:

1) Proposing a two-stage optimization formulation that guarantees desired coverage for computationally efficient structural inspection path planning.

2) Revealing the role and interaction of optimization parameters via a comprehensive sensitivity analysis to verify the applicability of the proposed method.

3) Demonstrating the flexibility of the proposed method for specialized inspection requirements such as focused inspection and no-fly zones.

## 2. METHOD OF STUDY

In the proposed path planning method, the input is a digital model from various sources, e.g., Computer-Aided Design (CAD) or a scanned 3D model generated using LiDAR or photogrammetry. Based on the model, the spatial configuration, including the viewpoint grid and visibility information are first generated to define the planning problem. Then, a two-step optimization process is performed. First, a genetic optimization algorithm is applied to identify the locations of points on the optimal path with the desired coverage and minimum length. The fundamental assumption at this first step is the presence of an omnidirectional camera at each point, leaving the camera pose estimation to the next step. Subsequently, a greedy algorithm is used to determine the camera poses of these viewpoints, given the angular specifications of the UAS camera. This formulation effectively splits the optimization problem into two smaller sub-problems. The process of the proposed path planning method is summarized in Fig. 1.

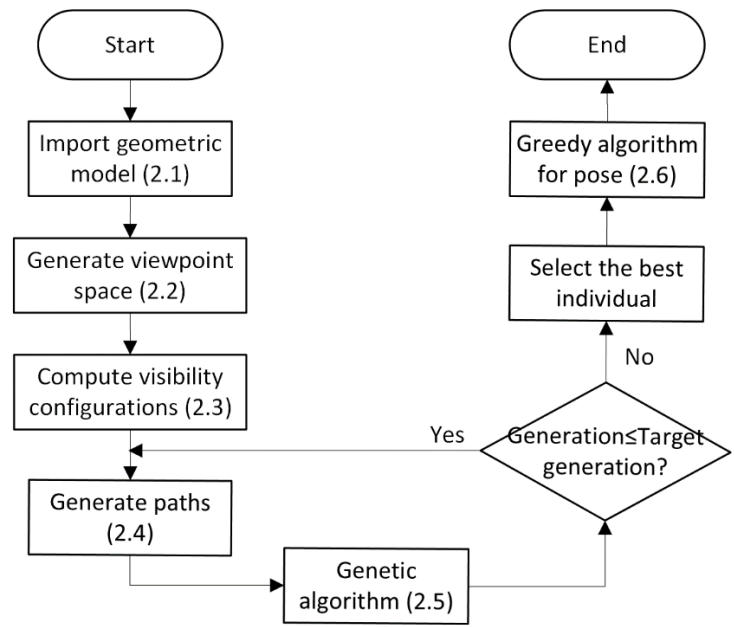

Fig. 1 Flowchart of the path planning method (the numbers refer to the corresponding section).

## 2.1 Modeling

The proposed framework is generic and compatible with various models, including CAD models, point clouds, and reconstructed meshes. The input model will be converted into a regular mesh for subsequent processing. The model is then divided into two parts: the inspection object and its surrounding environment (e.g., ground, trees, other obstacles, etc.). For visibility calculations, we exclusively use the mesh of the inspection object, while the surrounding environment is considered when determining the passable area of the UAS.

## 2.2 Grid-based graph configuration

To start the process, we discretize the space using a 3D grid and build a viewpoint space, which denotes the region around the object available for inspection. Next, we filter out the viewpoints inside the model (inspection object) or those with a distance to the model less than the safety threshold, addressing safety concerns. Based on the grid, a 26-connectivity 3D graph can be generated. For each vertex, there are at most 26 edges connecting it with its neighbor points, which have a distance no more than the grid size on all three axes, as shown in Fig. 2. The movement of the UAS will be limited in the viewpoint space and along the edges, which ensures that the path will be continuous without the UAS crashing into obstacles. An example of the generated viewpoint space is shown in Fig. 3.

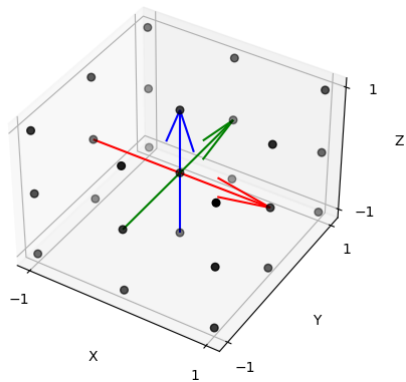

Fig. 2 26-connectivity graph for the vertex at the center.

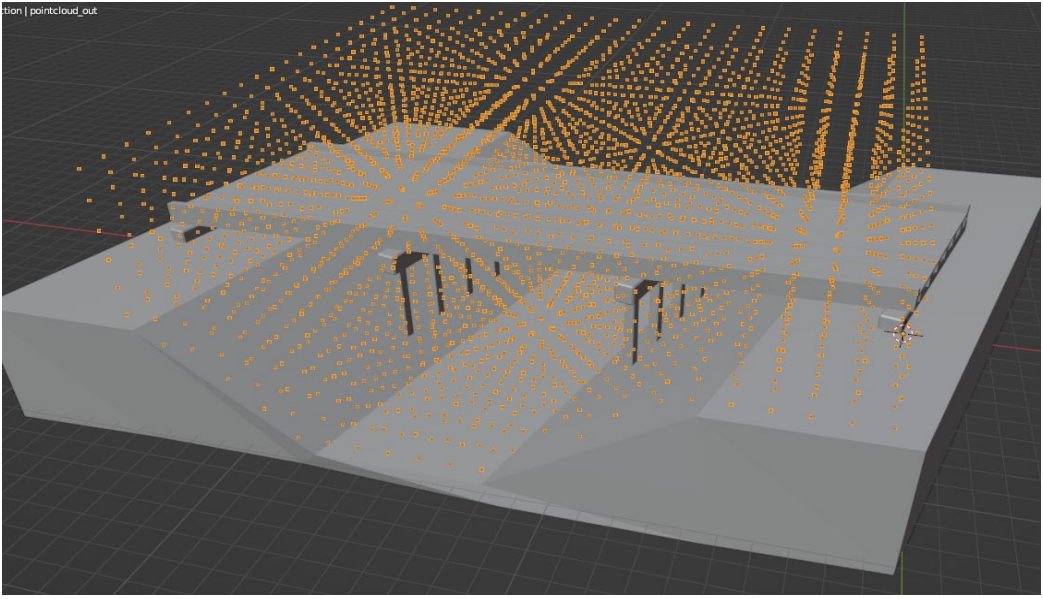

Fig. 3 Example of viewpoint space consisting of a regular grid with 1m spacing.

## 2.3 Visibility configurations

To ascertain the coverage of a path, it is essential to establish the visibility of the geometric model from the 3D viewpoint space through geometric relationships, as shown in Fig. 4. The process of determining visibility is explained in Algorithm 1. Firstly, the line segment connecting each viewpoint and the center of every face is generated, and the inclination angle between the line segment and the face is calculated. Subsequently, the presence of any intervening face between each viewpoint and the respective face is detected. By considering both the obstacle face and inclination angle, visibility information can be precomputed and stored in memory. This enables direct access to visibility during optimization instead of recalculating it in each optimization epoch. The coverage of a path can then be easily determined based on the visibility of each face.

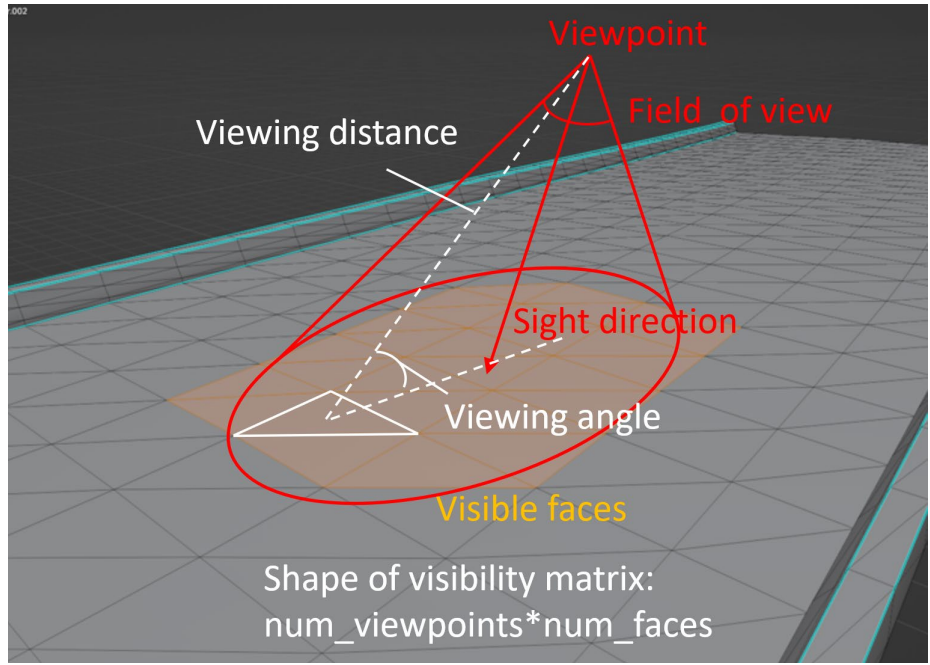

Fig. 4 Visibility of a face from a viewpoint.

Algorithm 1: Calculate visibility matrix

**Input:** $F\{f_j\}$: the collection of faces forming the bridge model
$VP\{vp_i\}$: the collection of all viewpoints
$vis\_dist$: distance threshold to determine visibility
$vis\_angle$: angle threshold to determine visibility

**Output:** $V\{v_{ij}\}$: visibility matrix indicating the visibility of each face from each viewpoint

1  **for** $vp_i$ *in* $VP$ **do**
     // loop for each viewpoint in the viewpoint space
2     **for** $f_j$ *in* $F$ **do**
     // loop for each face
3        $l \leftarrow vector(vp_i, f_j)$
4        **if** $angle(l, f_j) < vis\_angle$ *or* $distance(vp_i, f_j) > vis\_dist$
     // determine visibility based on angle and distance
5        **then**
6           $v_{ij} \leftarrow 0$
7        **else**
8           $v_{ij} \leftarrow 1$

**2.4 Path initialization**

For initialization, the path can be created by connecting a series of *n* consecutive (neighbor) viewpoints in the 3D space following the 26-connectivity graph shown in Fig. 2, where *n* denotes the number of viewpoints. Therefore, the positions of the viewpoints on the path can be encoded as an array with the shape of (*n*, 3), representing the path to be optimized. In this study, two types of initializations are defined: random initialization and rule-based initialization.

**Random initialization**

For random initialization, the choice of the viewpoints on the path is determined by randomly selecting an edge connecting to the current viewpoint, and the path is formed by iteratively moving forward until its length reaches the desired value. It is worth noting that the choice of points is restricted to the viewpoint space, eliminating the possibility of the UAS colliding with obstacles.

**Rule-based initialization**

Inspectors typically follow a rule-based progression of inspection tasks in manually controlled structural inspection tasks. For instance, they may fly the UAS along a grid path traversing every bridge span. While rule-based paths may not always be mathematically optimal, they are often reasonable starting points and can yield relatively good results. Moreover, once the rule is established, the paths can be automatically generated. It is hypothesized that applying rule-based initialization helps develop more realistic initial paths and improves the convergence speed. On the other hand, although most random initialized paths have poor performance compared to rule-based initialization, their higher flexibility contributes to inspecting difficult-to-reach areas with rule-based initialized paths. Therefore, combining these two types of initializations is expected to help balance exploration and exploitation in the optimization process.

In this study, the impact of such a rule-based initialization is studied by generating a portion of the initial paths using a rule-based strategy. The space around the bridge is divided into three spans along the bridge length, and several loops are generated for each span. Four points, two of which are above the deck, are randomly selected to form each loop. Fig. 5 provides a simple example path with one loop around each span to demonstrate rule-based initialization.

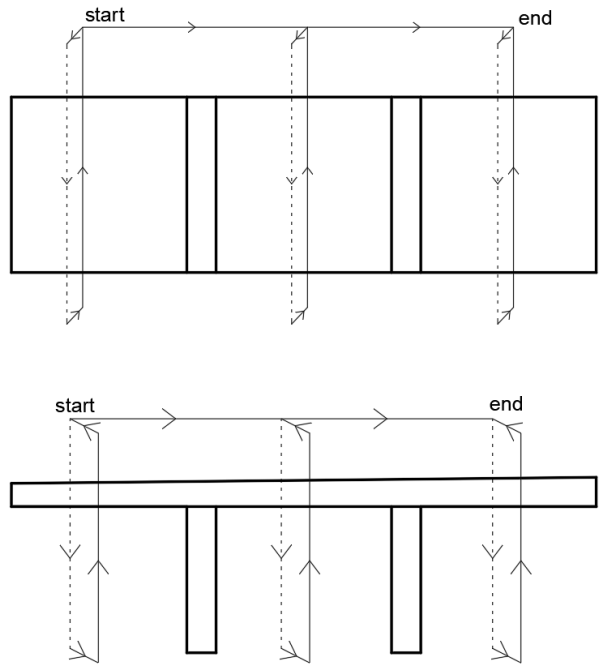

Fig. 5 A simple rule-based initialization path for demonstration.

## 2.5 Genetic optimization algorithm

To produce optimal paths capable of achieving the desired coverage while maintaining high efficiency, this study employs a genetic optimization algorithm (GA). This iterative approach is inspired by the process of natural selection and evolution and aims to progressively enhance the quality of the solution (in our case, the performance of the inspection path). As mentioned earlier, the path can be encoded into an array representing the positions of viewpoints, which is treated as the optimization variable. To ensure the completeness of the inspection, we propose to set coverage as a constraint that must exceed a threshold and regard path length as the optimization objective. In this way, a visibility path planning optimization problem is defined, whose solutions (i.e., inspection paths) are regarded as individuals that evolve with generations. In GA, generation refers to specific iterations or steps in the evolution process. In each generation, offspring are generated by applying mutation and crossover operators on the parents, which will be introduced in detail later. Individuals with more optimal fitness have a higher probability of surviving and passing their features to their offspring so that their performance can be iteratively improved.

**Overall optimization process**

Fig. 6 shows the process of the optimization algorithm applied to solve the path planning problem. Firstly, paths are initialized based on the two strategies discussed in 2.4. A selection operator is then employed to identify offspring exhibiting strong performance as parents for crossover and mutation operators. Performance in this context is characterized by a measure referred to as fitness. These operators are applied to generate offspring with new and improved attributes, offering the potential to surpass their parental counterparts. In each generation, the coordinated efforts of the mutation, crossover, and selection operators drive the iterative enhancement of individuals, thereby refining the coverage and efficiency of paths. The specifics of these operators are detailed in the next section.

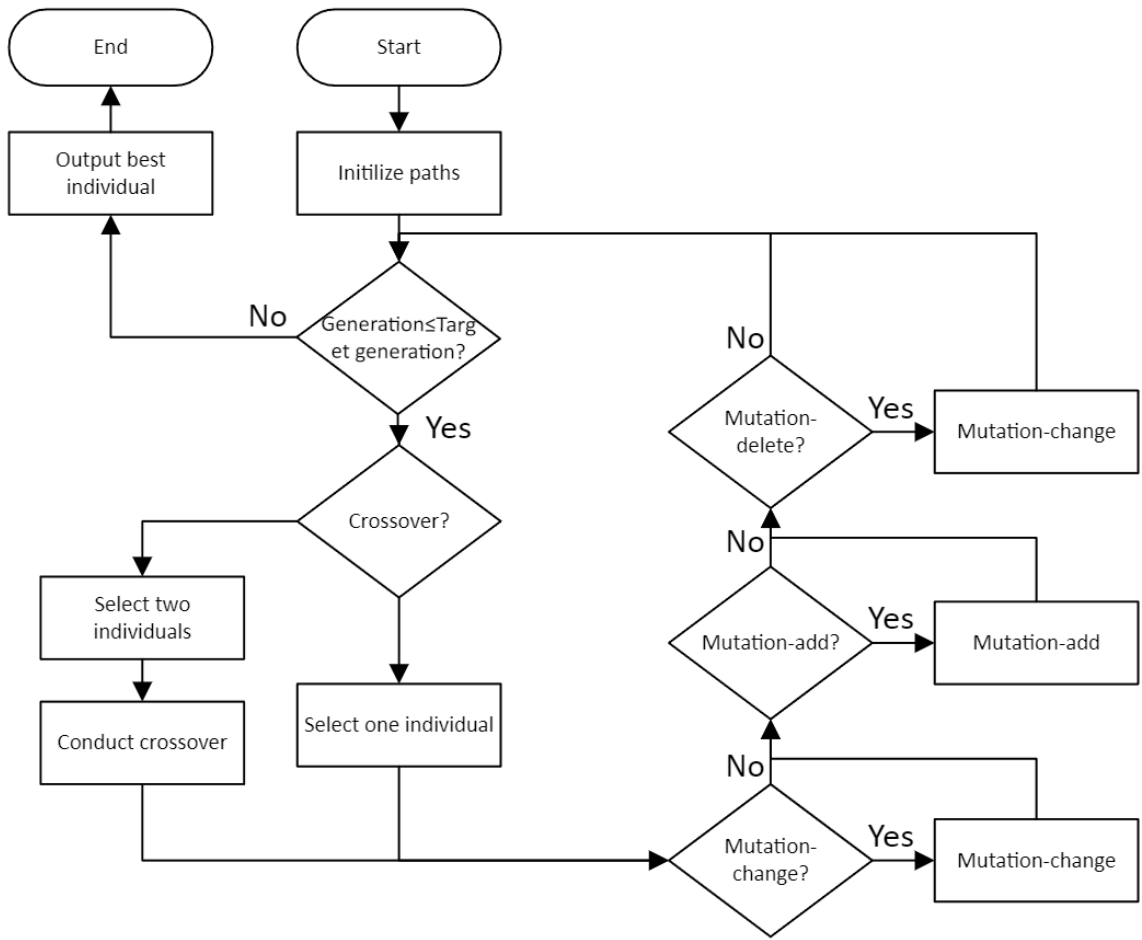

Fig. 6 Flowchart of the genetic optimization algorithm.

**Fitness**

In GA, fitness refers to a measure of how optimal a solution is or how well an individual performs with respect to the problem objectives. In general, we aim to achieve the shortest inspection path that produces visual coverage of the target structure. As we formulate the problem with coverage as a constraint, if the coverage of a path doesn't meet the constraint, improving coverage will be the primary objective. Otherwise, path length becomes the fitness value to be minimized. We define fitness as Eq. 1, where $\alpha$ is an extremely large value compared to path length. It is worth noting that fitness is always negative so that it can be maximized while path length is minimized.

$$Fitness = \begin{cases} -\dfrac{\alpha}{coverage} & \text{if } coverage < 0.95 \\ -path\_length & \text{if } coverage > 0.95 \end{cases} \qquad \text{(Eq. 1)}$$

In Eq. 1, the coverage is determined for each face based on Eq. 2

$$coverage = \dfrac{\sum_j \left[ (\sum_i v_{ij}^P) \geq w_j \right] * a_j}{\sum_j a_j} \qquad \text{(Eq. 2)}$$

in which $v_{ij}^P$ represents the visibility of face $j$ from point $i$ on the path. $w_j$ represents the weight of face $j$, which is always 1 for uniform weights. $a_j$ represents the area of face $j$. [] is the Iverson bracket, which returns to 1 if the statement inside is true and returns to 0 otherwise.

**Selection operator**

The selection operator determines which individuals from the current population will undergo further genetic operations. In this study, a tournament selector is used, wherein for each selection, a fixed number of individuals will be randomly selected from the parental generation to form a tournament, and the one with the highest fitness will be selected for the next generation. This method helps maintain diversity in the population while ensuring that individuals with better fitness have a higher chance of survival. The number of individuals selected for the tournament, which is denoted as tournament size, is the parameter that controls the balance between exploration and exploitation. If tournament size equals population size, the best population will always be selected, eliminating diversity. At the opposite extreme, if tournament size equals 1, the selection is totally random, thus curbing evolution and performance improvement. Therefore, a reasonable tournament size should be selected for better optimization convergence.

**Crossover operator**

Fig. 7(a) demonstrates the crossover operator. When performing a crossover operator, two paths are initially selected with the selection operator. We define two points as neighboring points if they are connected on the 26-connectivity graph, as shown in Fig. 2, or have the same coordinates. With this definition, all pairs of neighboring points are computed, for which the two points are on different paths. Subsequently, we assume the UAS starts from one path, and on each pair of neighboring points, it will have the probability of switching to the other path. This will enable random swaps to occur between two paths to generate offspring inheriting the advantages of both parents.

**Mutation operators**

Fig. 7(b) demonstrates three types of mutation operators, namely the change, add, and delete operators. The change operator alters the coordinates of viewpoints. When applied to a viewpoint, the viewpoints before and after on the path will be selected first. Then, it calculates the intersection of the two sets of neighbor viewpoints for these two points and randomly selects a new viewpoint from the intersection set to replace the original viewpoint. This ensures that the mutated path remains continuous. The add operator is similar to the change operator, with the key difference being that it examines the neighboring viewpoints of the current and next viewpoints. It randomly selects a point from the intersection set and inserts it after the current viewpoint. The delete operator creates new paths by removing points. Before deleting a viewpoint, it is necessary to verify whether the last viewpoint is among the neighboring viewpoints set of the next viewpoint to ensure continuity. If not, the delete operator won't be applied, ensuring continuity.

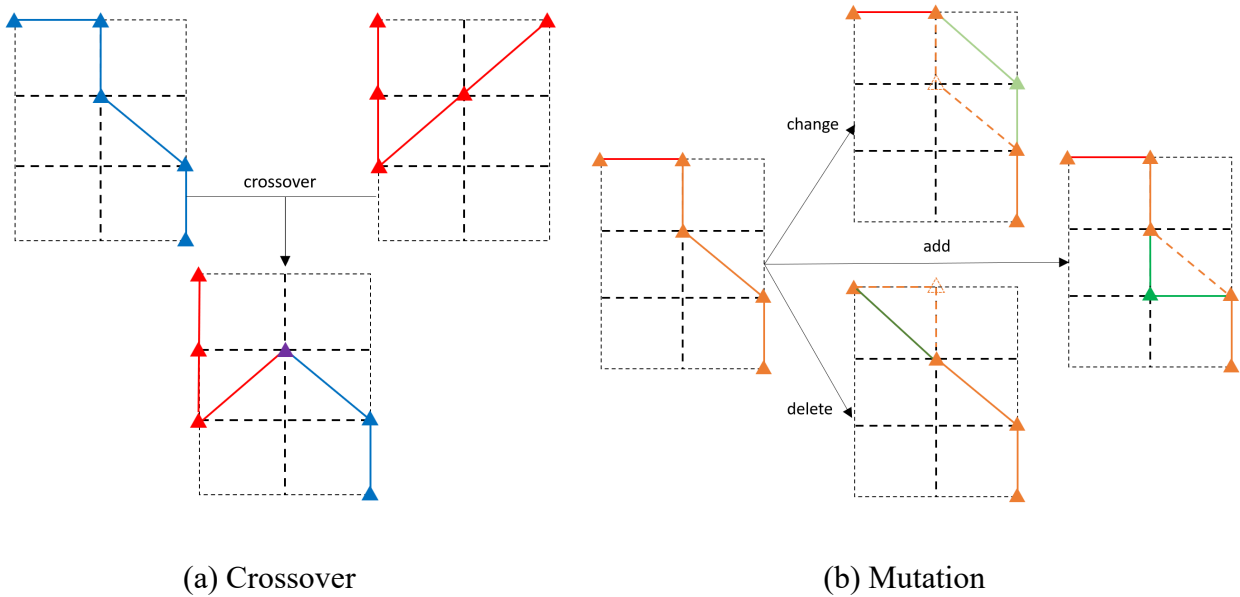

(a) Crossover          (b) Mutation

Fig. 7 Crossover and mutation operators.

A parameter, denoted as the individual evolution rate (IER), signifies the probability of initiating a mutation or crossover event for a path. If a crossover event is triggered, two individuals are selected from the parental generation, upon which the crossover operator is applied. Otherwise, the child will inherit attributes from only one individual from the parental generation. Then, a similar decision-making process occurs for the three mutation operators successively to determine if mutation will take place. For each crossover or mutation operator applied to an individual, a gene evolution rate (GER) is employed to determine the probability of applying the operator to each viewpoint for the mutation operator or pairs of neighbor viewpoints for the crossover operator.

**2.6 Greedy algorithm for pose calculation**

When determining the optimal path using GA, it is assumed that the UAS has a complete view of its surroundings, and the pose and FOV of the camera are not considered. Based on the viewpoint positions generated by the optimization, a series of pose vectors can be generated, pointing from each viewpoint on the path to each face. We then use a greedy algorithm to select the pose vectors to enable the poses to meet the coverage constraint, as explained in Algorithm 2. In essence, this algorithm iteratively selects the pose that covers the largest number of unseen faces until all the faces visible from the path are covered.

**Algorithm 2:** Determine the poses with greedy algorithm

**Input:** $F\{f_k\}$: the collection of faces forming the bridge model
$PV\{pv_i\}$: the collection of viewpoints on the path
$V\{v_{ij}\}$: visibility matrix
$FOV$: field of view angle

**Output:** $PP$: the collection of poses on the path

1 **for** $pv_i$ **in** $PV$ **do**
  // loop for each viewpoint in the path
2   **for** $ft_j$ **in** $F$ **do**
  // loop for each target face
3     **for** $fv_k$ **in** $F$ **do**
  // loop for each face to determine visibility
4       $sight_{ij} = vector(pv_i, ft_j)$
  // the sight from viewpoint to target face
5       $POSE \leftarrow POSE + (pv_i, sight_{ij})$
  // the pose including the information of viewpoint position and sight direction
6       $segment_{ik} = vector(pv_i, fv_k)$
  // the segment connecting each viewpoint and face to determine visibility
7       **if** $v_{ik} = 1$ and $angle(sight_{ij}, segment_{ik}) < FOV$ **then**
8         $v'_{ijk} \leftarrow 0$
9       **else**
10         $v'_{ijk} \leftarrow 1$
  // $v'$ is used to record if $f_k$ is visible from $pv_i$ with $sight_{ij}$

11 $FI \leftarrow F, n \leftarrow length(FI)$
  // initialization for invisible faces
12 **while** $n \neq 0$ **do**
13   **for** $pose_{ij} = (pv_i, sight_{ij})$ **in** $POSE$ **do**
14     $coverage_{ij} = \frac{\Sigma_{k=1}^{n} v'_{ijk} * area(f_k)}{\Sigma_{k=1}^{n} area(f_k)}$
  // calculate the coverage of each pose
15   $coverage_{i'j'} = max(coverage_{ij})$
  // pick the pose with maximum coverage
16   $FI \leftarrow FI - \{f_k | v'_{i'j'k} = 1\}$
  // delete the faces that are visible with the pose from the invisible faces collection
17   $PP \leftarrow PP + pose_{i'j'}$
  // record the poses with maximum coverage
18   $n \leftarrow length(FI)$
19 sort $PP$ so that the viewpoints positions have the same sequence as $PV$

## 3. EXPERIMENTAL SETUP

### 3.1 Model for experiment

To demonstrate the proposed mission planning algorithm, we use the model for an existing bridge in Champaign County, IL. Figure 8 shows the bridge, along with its geometric and mesh models. This bridge is a three-span structure with a total length of 25.3m, a skew of 15 degrees, and a structural system composed of reinforced concrete girders. The bridge carries two traffic lanes over a river with a vertical clearance of about 4 meters. Fig. 8 also depicts the geometric model constructed for the bridge and the mesh model generated with 3,369 triangular faces, each with an average size of $0.13m^2$.

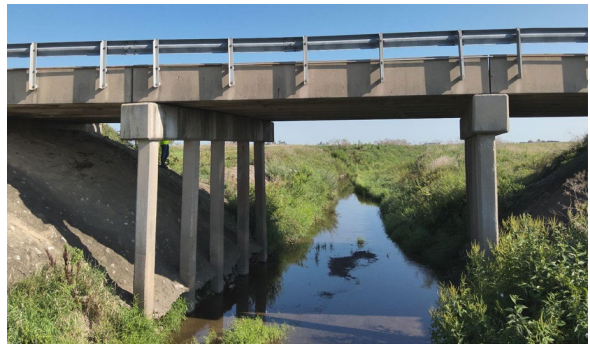

(a) Elevation view (on-site photo)

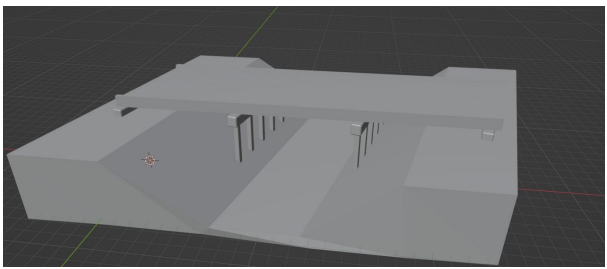

(b) Geometric model

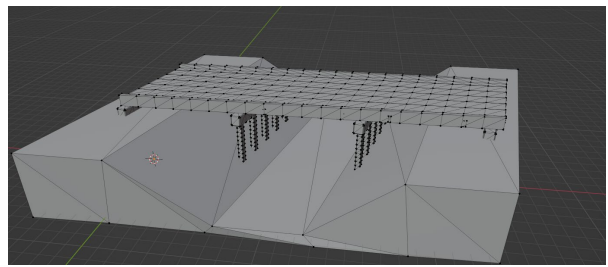

(c) Mesh model

Fig. 8 The bridge model used in this study.

## 3.2 Experimental parameters

This section provides a list of parameters and their definitions used in our experiments. The following four parameters determine the setup of the problem and are fixed throughout all our experiments. Table 1 summarizes the values chosen for these parameters.

**(1) Visible distance**: the distance threshold for determining if a face is visible from a viewpoint. Only when the distance between a face and a viewpoint is smaller than this threshold will the face be regarded visible.

**(2) Safety distance**: the threshold restricting the minimum distance between a viewpoint and a face to avoid accidents. All the viewpoints that have a distance to a face smaller than this value will be filtered out from the viewpoint space.

**(3) Visible inclination angle**: the angle threshold for determining if a face is visible from a viewpoint. Only when the angle between a face and the line of sight from a viewpoint to the face is smaller than this threshold will the face be regarded visible.

**(4) Grid interval**: the interval of the grid on which viewpoint space is generated.

Table 1 Parameters applied for all experiments.

| Grid interval | Visible distance | Safety distance | Visible inclination angle |
|---|---|---|---|
| 1m | 10m | 0.5m | 45° |

The following parameters affect the optimization algorithm and are varied across their respective practical ranges throughout our sensitivity analyses.

**(1) Individual evolution rate (IER):** the probability of a child path going through mutation or crossover.

**(2) Gene evolution rate (GER):** the probability for mutation to happen on each viewpoint, and for crossover to happen between two pairs of neighbor points.

**(3) Tournament size:** the number of individuals randomly selected for each tournament selection.

**(4) Rule-based initialization proportion:** the proportion of initial paths generated by rule-based initialization. The remaining paths are generated following random initialization.

**(6) Coverage goal:** the minimum coverage requirement for a path.

**(7) FOV:** the field of view of the camera. This parameter will only influence the results of the greedy algorithm.

## 3.3 Sensitivity analysis settings

We conducted a parametric analysis to research the influence of the parameters on the optimization results. Table 2 lists the values used for comparison in our sensitivity analysis, and the highlighted numbers are the parameters used for the baseline case. For each parameter setting, ten repetitions of the tests are conducted to quantify the impact of randomness in the optimization process.

Table 2 Parameters for the baseline case.

| IER | GER | Tournament size | Rule-based initialization proportion |
|---|---|---|---|
| 0.5 **0.75** 1 | 0.05 **0.1** 0.2 0.5 | 12 **25** 50 | 0 **0.5** 1 |
| Population size | Generations | Coverage | FOV |
| 125 **250** 500 | **500** | 90% **95%** 99% | 60° **90°** 120° |

## 4 RESULTS AND DISCUSSION

### 4.1 Feasibility analysis

As stated before, using random initial paths and random operators such as mutation, crossover, and selection results in variations in the optimized response. We repeat each test ten times to demonstrate and quantify this randomness. The evolution of the ten experiments for the baseline case is shown in Fig. 9. Overall, all of the experiments successfully achieve the desired 95% coverage goal, and path length consistently decreases with generation and eventually converges to a plateau. The average optimal path length is 162.2m, and the standard deviation is 5.8m, indicating that the optimization algorithm is robust despite its random nature. An interesting fact is that, for the second test, detrimental mutation or crossover happened before the 170th generation, leading to an increase in path length. However, the optimization gets back on track at around the 200th generation and reaches a level comparable to other tests, which indicates that the genetic algorithm successfully corrected the harmful mutations. These results reveal that the optimization algorithm demonstrates reliable convergence to a steady level.

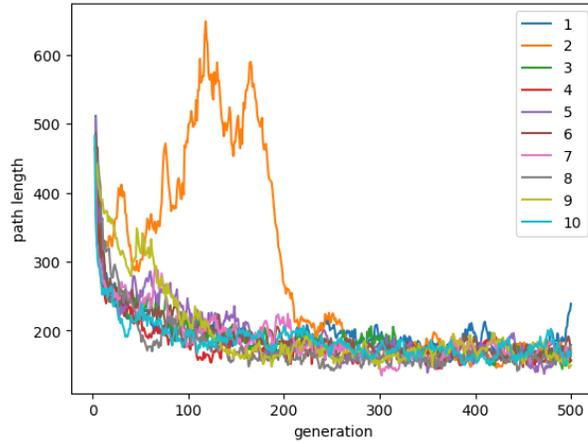

Fig. 9 The evolution for ten tests using base case parameters.

Fig. 10 shows one of the optimized paths and the corresponding visible surfaces. The optimized paths tend to wind around the entire model, take pictures from different directions and avoid repetitions to ensure efficiency, which intuitively demonstrates the overall success of the proposed SIPP method.

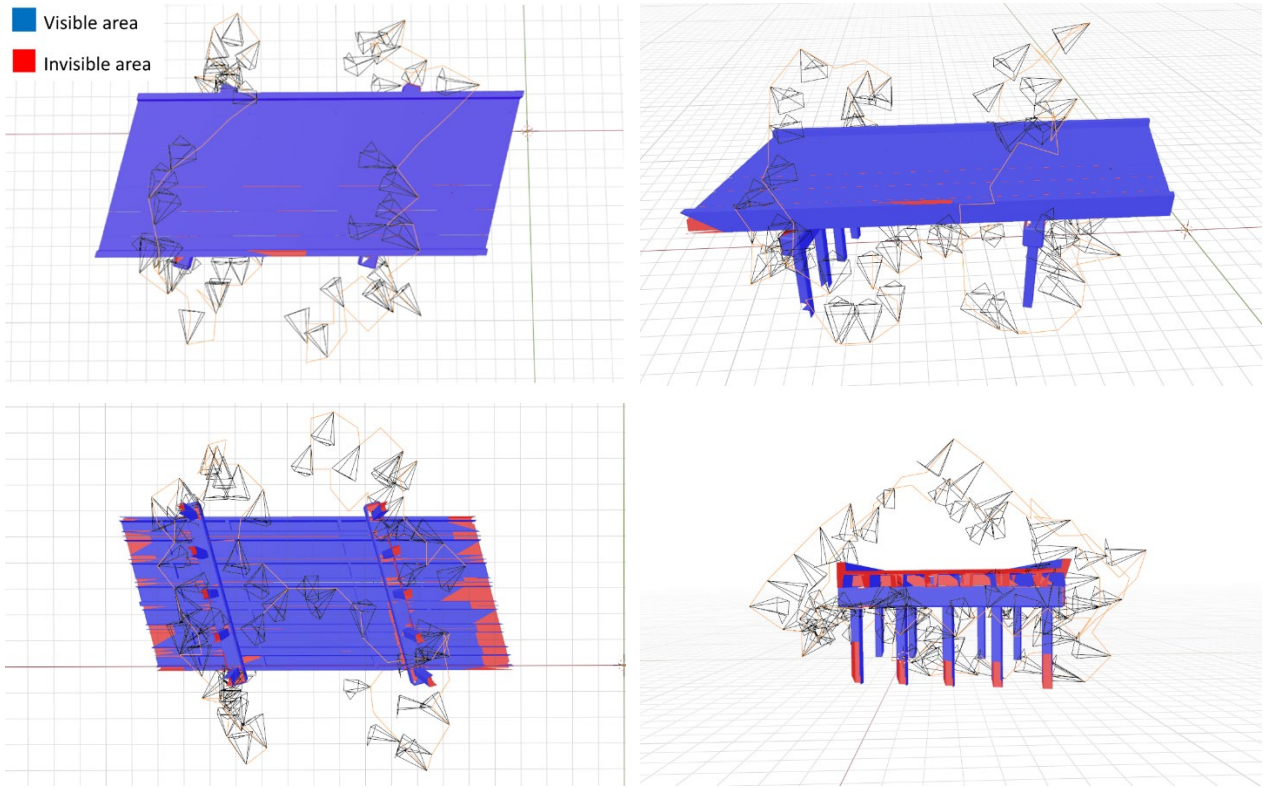

Fig. 10 The generated optimal path and visible surface (95% coverage) viewed from different directions.

Fig. 11 provides a more exhaustive analysis of the efficiency of the optimized path. Fig. 11(a) shows the increasing trend of the cumulative coverage with the addition of new viewpoints on the path during the inspection process, which has a steady upward trend. This shows that most of the viewpoints on the inspection path effectively contribute to generating visual coverage of the bridge. Fig. 11(b) shows the total coverage of each viewpoint (independent of other viewpoints) and the exclusive contribution of each viewpoint to coverage (the faces that are already visible from previous viewpoints are not counted). It can be seen that most viewpoints are uniquely contributing to coverage, while a few viewpoints provide duplicate coverage. The reason is that the path needs to be continuous, resulting in some inefficient viewpoints that primarily serve as the connection between other points.

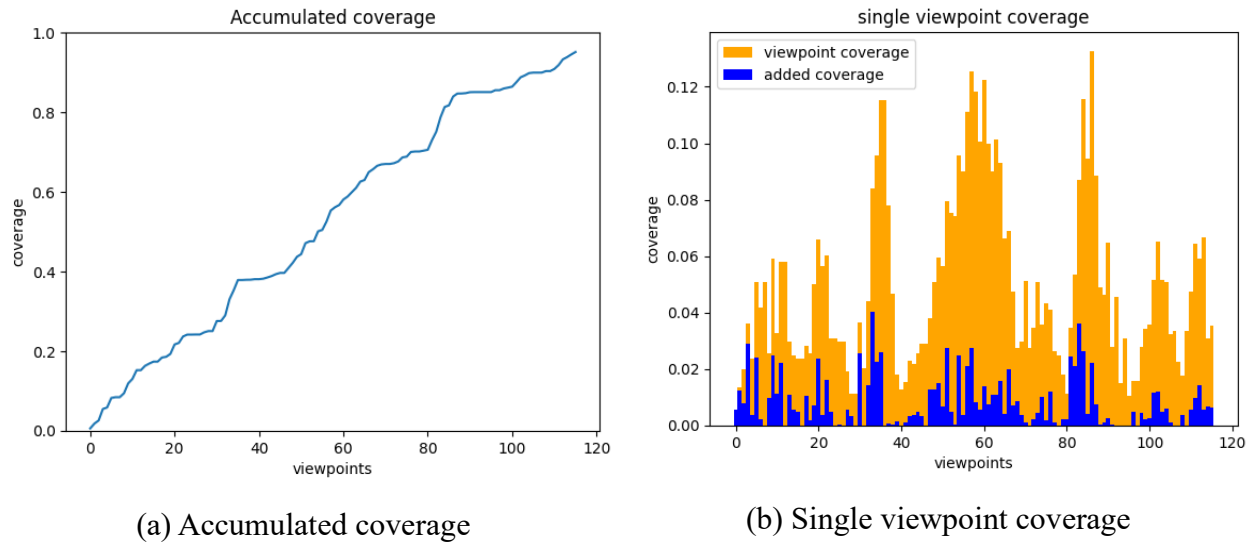

(a) Accumulated coverage    (b) Single viewpoint coverage

Fig. 11 The contribution of viewpoints on the optimal path to coverage.

### 4.2 Influence of population and generation

Fig. 12 illustrates the evolutionary process for various population sizes. Across all three population settings, path length stabilizes after approximately 300 generations, indicating that a maximum generation limit of 500 is sufficient for optimization convergence. A comparison between population settings of 125 and 250 reveals that an inadequate population size results in suboptimal optimization outputs. Moreover, while a larger population of 500 initially leads to faster convergence than 250, both eventually reach a similar level after 500 generations. Considering the additional computational expense of higher population size and maximum generation settings, careful selection of parameters is crucial to achieve a balance between performance and computational cost.

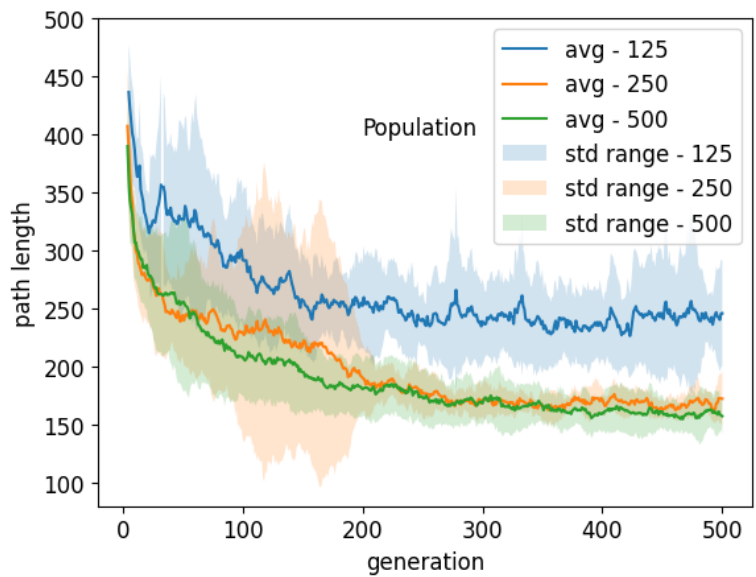

Fig. 12 The influence of population size and generation on the optimization process.

## 4.3 Influence of evolution rates

We conducted experiments for different combinations of IER and GER, and the average and standard deviation of optimal path lengths are summarized in Fig. 13. It is shown in Figure 13 that both IER and GER have a significant influence on optimization results. It can be observed that an intermediate value for the two evolution rates leads to the most optimal paths, and the extreme values yield worse results. The reason is that the evolution rate controls the balance between exploration and exploitation. A higher evolution rate leads to more exploration of the solution space, which means that there is a greater chance of introducing random changes to the individuals, potentially enabling the algorithm to discover new, unexplored regions. However, excessive exploration may hinder the algorithm's ability to exploit the promising solutions it has already found. Additionally, with a low GER, higher IER will lead to better results, indicating that the GER and IER tend to compensate each other. Although extreme values are undesirable, the optimization results demonstrate robustness within a certain range of evolution rates. This suggests that, despite the influence of evolution rates on the results, the optimization process is not overly sensitive to these parameters.

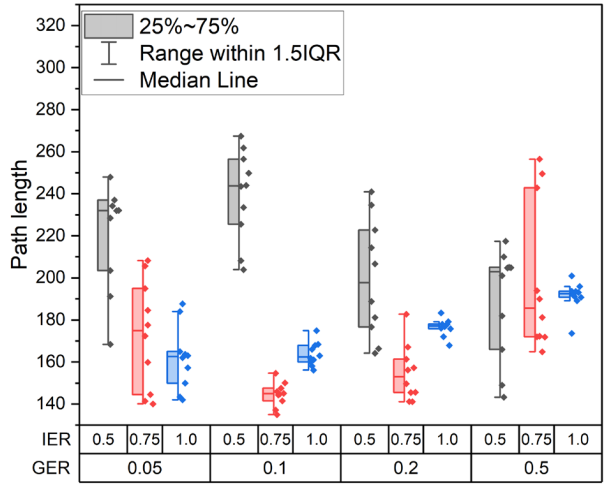

Fig. 13 The effect of evolution rates on path length.

## 4.4 Influence of rule-based initialization proportion

We conducted experiments for different rule-based initialization proportions, and the results are shown in Fig. 14. This figure shows that, compared with a totally random initialization strategy, using rule-based initialization significantly decreases the path length in the first 100 generations, which indicates that rule-based initialization helps to accelerate the convergence of the optimization. Although the difference decreases with evolution, combining rule-based initialization with random initialization still shows apparent advantages in the final optimal path length. When all the paths are initialized based on rules, there is no significant difference for the first 150 generations, while it performs worse than the baseline case eventually. The reason is that random initialization provides paths with more flexibility, which balances the drawback of rule-based initialization.

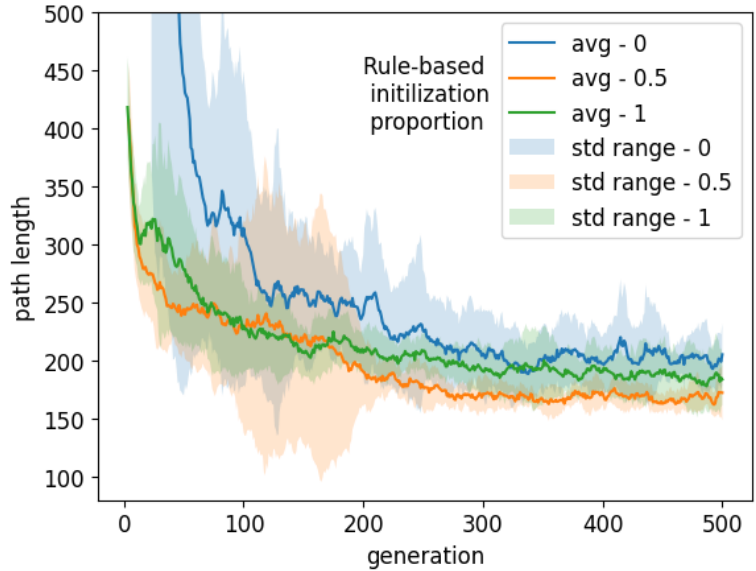

Fig. 14 The influence of rule-based initialization proportion on the optimization process.

**4.5 Influence of tournament size**

We also examined the impact of tournament size on optimization convergence, as presented in Fig. 15. Similar to the results observed for evolution rates, the shortest path length is achieved with a moderate tournament size of 25. Using a larger tournament size of 50 increases the path length by approximately 19%, while a smaller tournament size of 12 results in about a 12% increase in path length. Therefore, selecting an appropriate tournament size can enhance optimization results, although values that are excessively large or small can still produce acceptable outcomes.

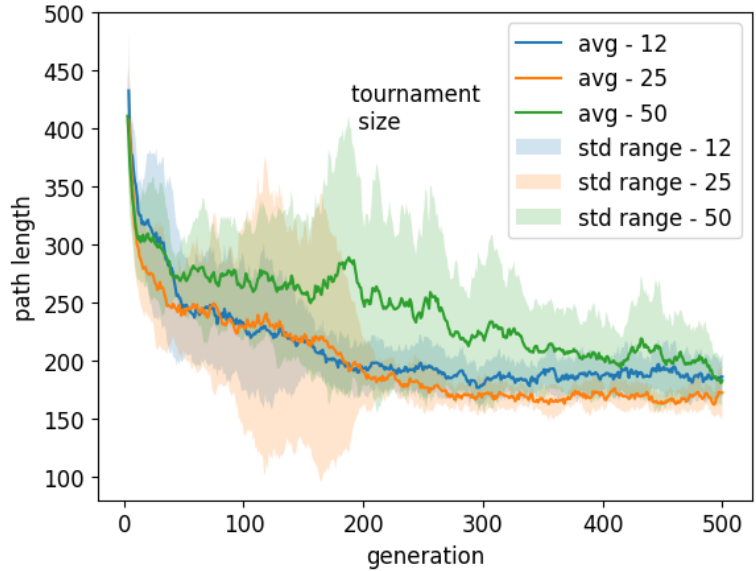

Fig. 15 The influence of tournament size on the optimization process.

## 4.6 Influence of FOV

In the proposed two-step formulation, FOV does not influence the GA optimization process (due to the omnidirectional camera assumption) and only affects the greedy pose determination step. We investigated the relationship between FOV and the number of poses required to fulfill the coverage constraint in the baseline case. As shown in Fig. 16, a larger FOV reduces the number of poses needed, suggesting that using a camera with a larger FOV can enhance inspection efficiency. This is because a larger FOV provides higher visibility from each viewpoint, similar to using a wide-angle camera. However, it is worth noting that, in practice, an excessively large FOV can lead to distortion and adversely impact inspection quality.

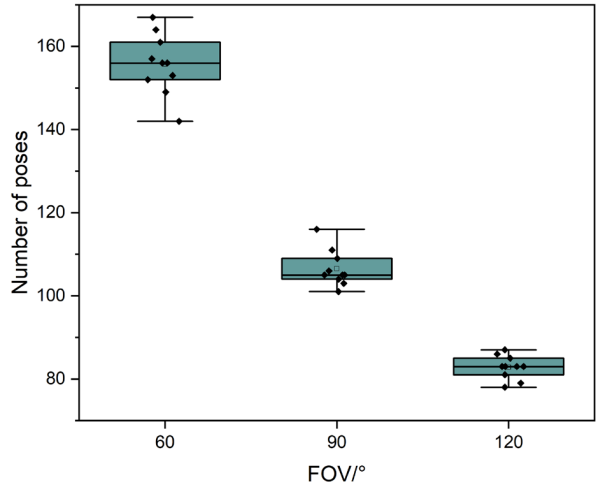

Fig. 16 The distribution of the number of poses needed for different FOV values.

**4.7 Influence of coverage constraint**

The coverage constraint is a crucial factor in optimization. As illustrated in Fig. 17, when the coverage constraint is raised from 0.9 to 0.95, the path length increases gradually with minimal variation, signifying the method's reliability. With the stringent 0.99 coverage constraint, there is a substantial increase in both path length and standard deviation, indicating that higher coverage demands make convergence more challenging while all experiments still successfully converge.

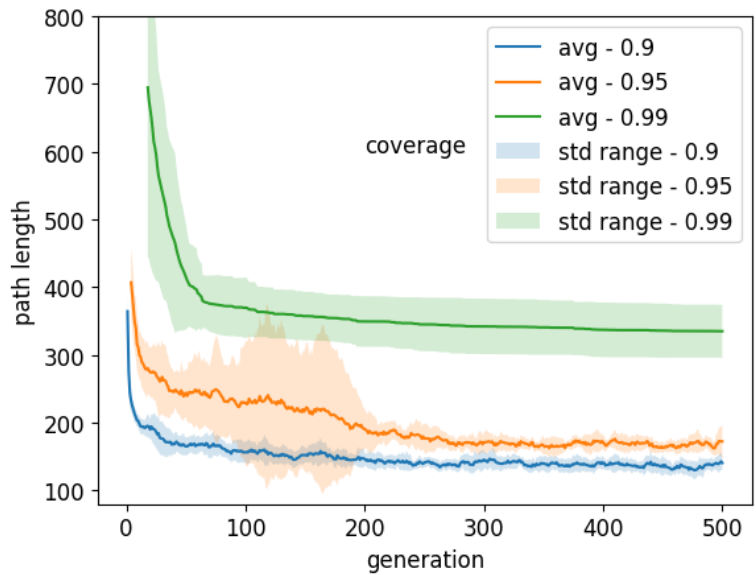

Fig. 17 Influence of coverage constraint on the optimization process.

## 4.8 Partial space inspection and no-fly zones

In real-world applications, various restrictions are often in place for safety and practical reasons. For instance, flying a UAS over a bridge when traffic is not restricted poses a risk of driver distraction and accidents. In fact, flight above the deck may be unnecessary when the goal is the structural inspection of the substructure and superstructure, and the cracks on the roadway are not of immediate interest. Given the complexity of practical engineering scenarios, enabling partial space inspection is crucial for effective and practical structural inspection mission planning. In our proposed method, achieving this is conveniently accomplished by modifying the definition of the viewpoint configuration space and defining no-fly zones in areas of interest. The results of restricting UAS flight above the bridge are illustrated in Fig. 18. In this figure, the UAS moves under the bridge and covers everywhere except the road above, thus avoiding driver distraction while inspecting the substructure and superstructure.

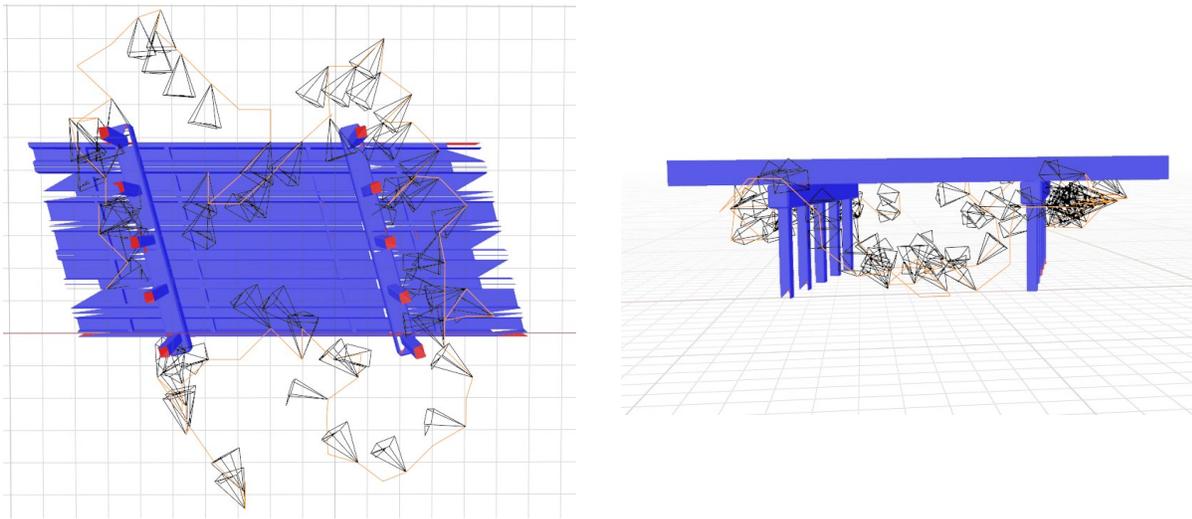

Fig. 18 Visualization for partial space inspection paths (no-fly zone above the bridge).

## 4.9 Focused inspection

Different parts of a structure may have different degrees of criticality to the overall performance and safety and may be subject to different levels of load-carrying demand and deterioration. It follows that effective inspections should take these disparities into account. As described earlier in this paper, the proposed framework allows for the assignment of varying degrees of importance to different parts of the structure by means of different weights for different structural components. To demonstrate, assume the case where the bottom of the girders at mid-span and the bridge deck joints above columns are identified as parts where damage tends to appear. To ensure higher attention to the more critical zones, a weight term ($w_j$) of 2 is assigned to them in Eq. 2 as opposed to a weight of 1 everywhere else, thus guaranteeing that these faces will be scanned at least twice as many times as the rest of the structure. The critical zones are colored red in Fig. 18 (a) and (b). Fig. 18 (c) and (d) show the resulting heatmap of the frequency of faces being visited, and Fig. 18 (e) and (f) show the resulting heatmap with uniform weight as comparison. It can be observed that when weights are assigned, the critical parts are visited more frequently compared with adjacent areas. This example demonstrates the capability of the proposed methodology to prioritize critical details and components that require more stringent inspections based on prior knowledge from engineering expertise, structural analysis, or history of deteriorations.

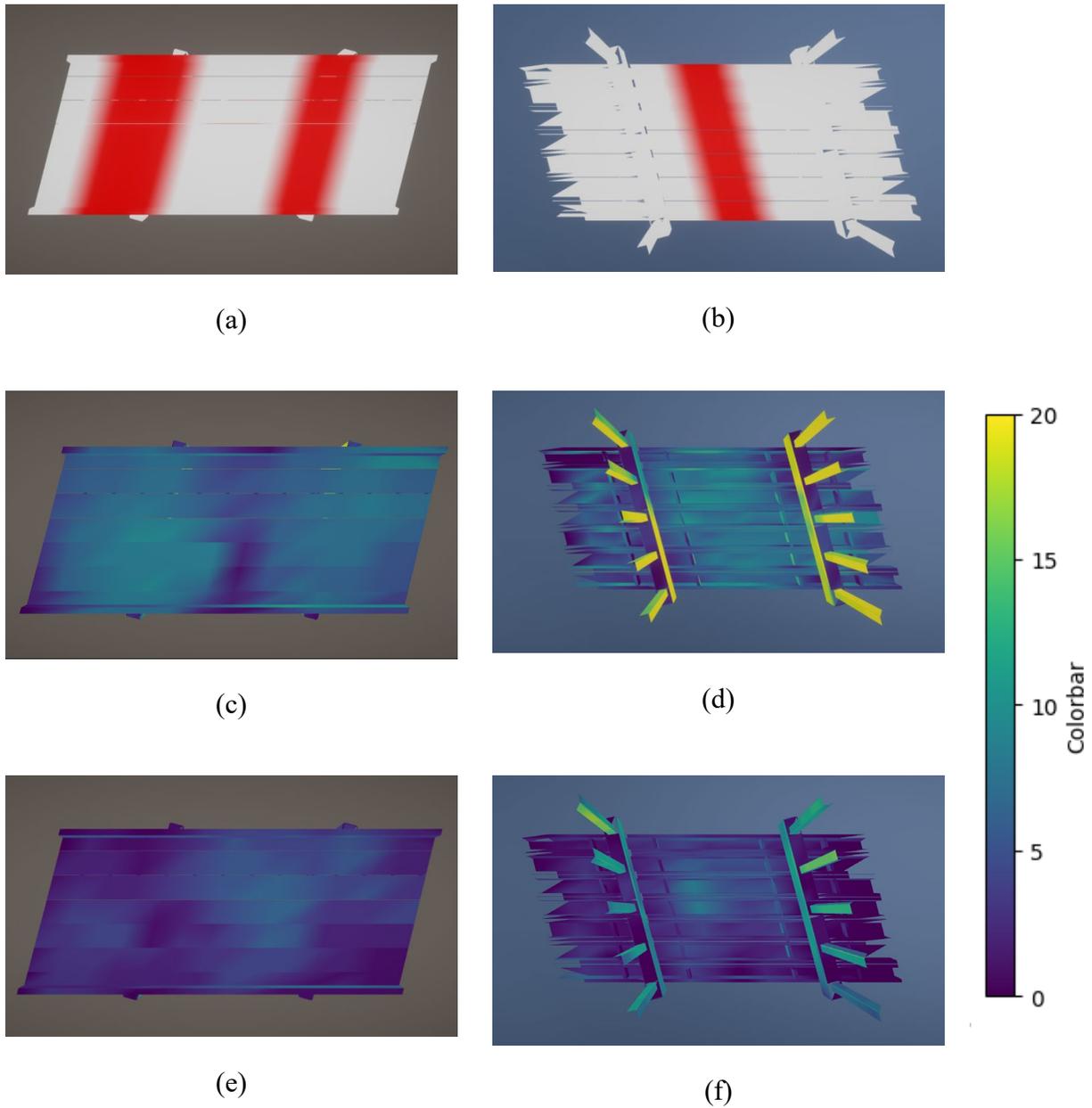

Fig. 19. Heatmap of faces being visited with weighted overlap inspection strategy.

**CONCLUSIONS**

In this study, a comprehensive pipeline for UAS-based structural inspection path planning was proposed to enhance the efficiency and reliability of visual data collection. The proposed SIPP process encompasses several key stages, including modeling, space configuration, visibility quantification, path optimization, and greedy pose determination. To illustrate the proposed methodology, we employ a 3-span concrete girder bridge as a demonstrative example. The feasibility of employing a genetic optimization algorithm to solve SIPP was demonstrated by the consistent success of the technique in converging to paths with the desired coverage constraint. Visualization and analysis of the resulting paths reveal their tendency to meander around the bridge, with most viewpoints uniquely contributing to coverage accumulation, ensuring that the coverage constraint is met efficiently. Notably, the optimization algorithm demonstrates automatic error correction capabilities when detrimental mutations occur.

The sensitivity analysis of the optimization parameters highlighted the significant influence of reasonable population and generation settings on convergence and computational cost-effectiveness. Moreover, it was demonstrated that evolution rates can be fine-tuned to improve the quality of optimal results. The application of partial rule-based initialization and a moderate tournament size also contribute to enhancing optimization outcomes. It was further demonstrated that the inspection parameters, including coverage constraint and FOV have a significant impact on the optimization and greedy pose determination. Notably, higher coverage constraints and smaller FOV values result in more intricate inspection paths and poses, aligning with intuitive expectations.

At last, we explored the potential of the proposed method to address specific SIPP characteristics and demonstrate its flexibility and broad applicability. In one case, we limit the inspection space to the underside of the bridge to prevent distractions to the drivers above. In another, we assign different importance levels to structural components based on engineering knowledge, introducing a weighted overlap requirement for optimization. This enables inspectors to assign greater importance to critical components, thereby enhancing the overall efficiency of inspection.

While the proposed methodology successfully achieved the desired objectives, we recognize opportunities for further refinement. As the current method relies on a model as input, the application of a more advanced algorithm, such as reinforcement learning, holds promise for real-time online path planning, which would not rely on a pre-existing model of the structure. Additionally, considering additional performance metrics, such as 3D model generation and visual detection quality, can further enhance the capacity of the proposed method to meet the diverse demands of SIPP. Given the growing use of UAS, we believe that the proposed method lays a solid foundation for implementing automated UAS-based inspection approaches in infrastructure inspection.

ACKNOWLEDGEMENTS

The authors would like to express their sincere appreciation to the NVIDIA Corporation for their generous support with the donation of GPUs used in this study.